\documentclass[letterpaper, 10 pt, conference]{ieeeconf}  

\IEEEoverridecommandlockouts                              
\usepackage[utf8]{inputenc}
\usepackage{amsmath,amssymb,stmaryrd,mathtools}
\usepackage{amsfonts}
\usepackage[T1]{fontenc}
\usepackage[noend]{algpseudocode}
\usepackage{acronym}
\usepackage{xcolor}
\usepackage{verbatim}
\usepackage{booktabs}
\usepackage{siunitx}
\usepackage{graphics}
\usepackage{graphicx, caption}
\usepackage{flushend}
\usepackage{suffix}
\usepackage{xstring}
\usepackage{xparse}
\usepackage{expl3}
\usepackage{mathrsfs}
\usepackage{tabularx}
\usepackage{makecell}
\usepackage{array}
\usepackage{hyperref}

\usepackage{cleveref}
\usepackage[ruled,linesnumbered]{algorithm2e}
\usepackage{multirow}
\usepackage{diagbox}
\usepackage{rotating}

\usepackage{subcaption}
\usepackage{wrapfig}

\usepackage{tikz}
\usetikzlibrary{arrows,backgrounds,calc}
\usepackage{relsize}
\usepackage{float}
\usepackage{kantlipsum} 
\usepackage{lipsum}
\usepackage{stfloats}
\usepackage{siunitx}

\usepackage{xcolor}

\usepackage[noadjust]{cite}
\usepackage{url}

\usepackage{todonotes}
\usepackage{soul}
\definecolor{smoothgreen}{rgb}{0.7,1,0.7}
\sethlcolor{smoothgreen}

\RequirePackage{luatex85}
\usepackage{pgfplots}
\pgfplotsset{compat=newest}
\pgfplotsset{every axis legend/.append style={%
		cells={anchor=west}}
}
\usepgfplotslibrary{polar}
\usetikzlibrary{arrows}
\tikzset{>=stealth'}

\definecolor{C1}{rgb}{0.0, 0.447, 0.741}
\definecolor{C1_light}{rgb}{0.0, 0.6032388663967612, 1.0}
\definecolor{C2}{rgb}{0.85, 0.325, 0.098}
\definecolor{C3}{rgb}{0.929, 0.694, 0.125}
\definecolor{C4}{rgb}{0.494, 0.184, 0.556}
\definecolor{C5}{rgb}{0.466, 0.674, 0.188}
\definecolor{C6}{rgb}{0.301, 0.745, 0.933}
\definecolor{C7}{rgb}{0.635, 0.078, 0.184}

\usepgfplotslibrary{groupplots}

\usetikzlibrary{shapes.geometric, arrows}

\tikzstyle{startstop} = [rectangle, rounded corners, minimum width=2cm, minimum height=1cm,text centered, draw=black, fill=none]
\tikzstyle{arrow} = [thick,->,>=stealth]

\renewcommand{\baselinestretch}{0.915} 

\title{
Learning Coordinated Bimanual Manipulation Policies \\ using State Diffusion and Inverse Dynamics Models
}

\author{Haonan Chen, Jiaming Xu\textsuperscript{*}, Lily Sheng\textsuperscript{*}, Tianchen Ji, Shuijing Liu, Yunzhu Li,  Katherine Driggs-Campbell%
\thanks{H. Chen, J. Xu, L. Sheng, T. Ji, and K. Driggs-Campbell are with the University of Illinois at Urbana-Champaign. S. Liu is with the University of Texas at Austin. Y. Li is with Columbia University. Emails: \{haonan2, jx30, lilys3, tj12, krdc\}@illinois.edu, shuijing.liu@utexas.edu, yunzhu.li@columbia.edu.\newline
* Equal contribution. }
}

\begin{document}
\maketitle

  

\begin{abstract}
When performing tasks like laundry, humans naturally coordinate both hands to manipulate objects and anticipate how their actions will change the state of the clothes. However, achieving such coordination in robotics remains challenging due to the need to model object movement, predict future states, and generate precise bimanual actions.
In this work, we address these challenges by infusing the predictive nature of human manipulation strategies into robot imitation learning. Specifically, we disentangle task-related state transitions from agent-specific inverse dynamics modeling to enable effective bimanual coordination. Using a demonstration dataset, we train a diffusion model to predict future states given historical observations, envisioning how the scene evolves. Then, we use an inverse dynamics model to compute robot actions that achieve the predicted states.
Our key insight is that modeling object movement can help learning policies for bimanual coordination manipulation tasks.
Evaluating our framework across diverse simulation and real-world manipulation setups, including multimodal goal configurations, bimanual manipulation, deformable objects, and multi-object setups, we find that it consistently outperforms state-of-the-art state-to-action mapping policies. Our method demonstrates a remarkable capacity to navigate multimodal goal configurations and action distributions, maintain stability across different control modes, and synthesize a broader range of behaviors than those present in the demonstration dataset. 
\end{abstract}


\section{Introduction}
\label{sec:intro}
\begin{figure*}[tbp]
  \centering
  \includegraphics[width=0.84\linewidth]{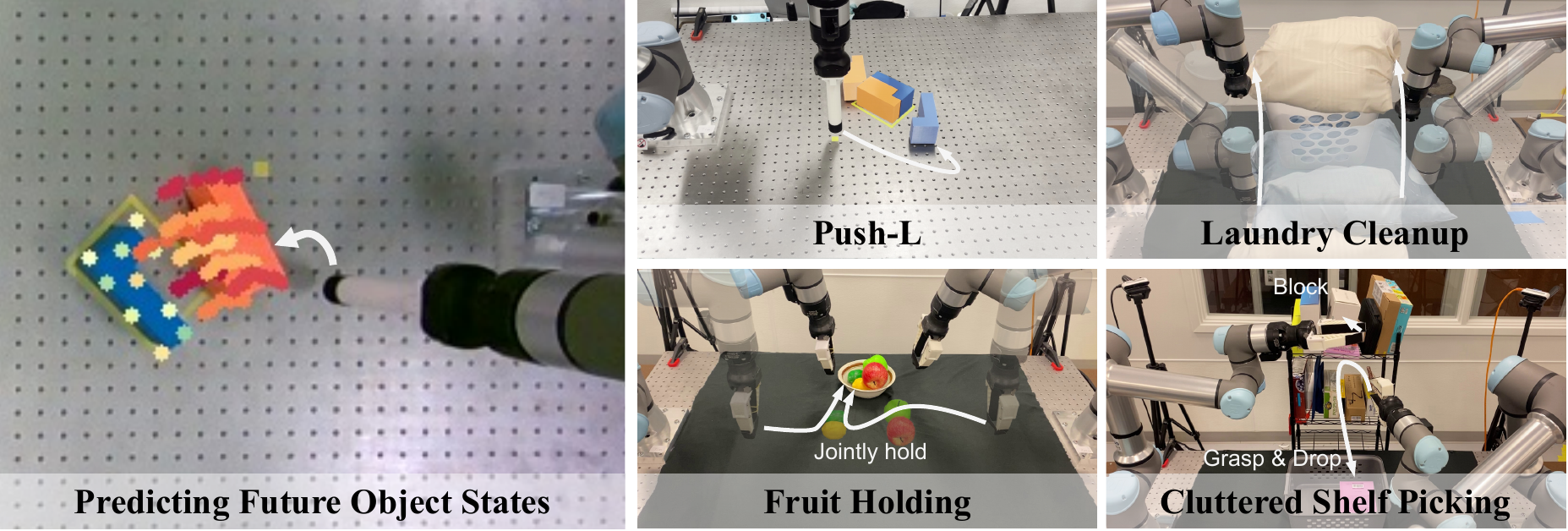}
  \caption{\textbf{Prediction-aided imitation learning for coordinated bimanual manipulation.} In the left image, the L-shaped blocks are represented by keypoints, with their predicted future trajectories visualized. The diffusion model predicts future states, which the inverse dynamics model uses along with the previous state to generate actions. Our framework is validated with underactuated systems, deformable objects, bimanual coordination, and multi-object interactions, demonstrated in tasks such as push-L, laundry cleanup, fruit holding, and cluttered shelf picking. \vspace{-5pt}}
  \label{fig:teaser}
  \vspace{-10pt}
\end{figure*}
{Many everyday bimanual manipulation tasks, such as cooking or sorting laundry, are simple for humans but remain challenging for robots. Humans naturally anticipate how their actions will influence object states, using predictive reasoning to guide movements~\cite{TEGLAS2016227, https://doi.org/10.1111/desc.13042}. Unlike single-arm tasks, which primarily involve independent end-effectors, bimanual tasks demand cooperative force distribution, complex spatial planning, and interaction-aware control, making it difficult for robots to achieve stability and precision, especially in tasks involving deformable or multiple objects.
}




{Despite recent advances in robotic manipulation~\cite{chen2023predicting, doi:10.1177/02783649231219020, liu2022intention, shi2023robocook}, bimanual coordination remains an open challenge due to the intricate interplay between robot actions and object dynamics. Many imitation learning methods rely on end-to-end state-action mapping, which struggles to generalize across multimodal goal configurations and unseen interactions~\cite{ibc,bet}. In contrast, humans excel at using both hands simultaneously because they explicitly anticipate object movement before executing actions. This predictive strategy in human manipulation motivates our approach to incorporating state prediction into imitation learning.
}

To overcome these limitations, we propose a state prediction-aided imitation learning framework that explicitly models future states and actions, which enhances spatial and force reasoning, as well as interaction-aware control for bimanual coordination. 
By leveraging diffusion models, which have shown great performance in image, video generation, and trajectory synthesis~\cite{rombach2022high, https://doi.org/10.48550/arxiv.2204.11824, ho2022video, blattmann2023videoldm, Janner2022PlanningWD, ajay2023is}, we improve the robot's ability to anticipate future states and coordinate robot actions effectively (Figure~\ref{fig:teaser}).
Our approach integrates a diffusion model to predict future states from historical data and an inverse dynamics model to generate necessary actions to achieve these predicted states.


{Our key insight is that explicitly modeling object movements greatly aids bimanual coordination. When two robots coordinate to manipulate an object, such as lifting or moving a large or deformable item, a failure like dropping the object leads to a large state loss in our model, highlighting coordination failure. In contrast, state-to-action mapping approaches may not show significant action loss for similar robot trajectories with an object drop, masking critical coordination issues. Modeling the object state allows our method to capture these errors more effectively, resulting in better control in complex bimanual tasks.}

Our contributions are summarized as follows: 
{(1) We propose a novel imitation learning framework that separates state prediction from inverse dynamics modeling, improving long-term planning and bimanual coordination.
(2) We conduct a comprehensive studies, demonstrating the necessity of both the diffusion model and inverse dynamics model for complex, coordinated bimanul manipulation tasks.}
(3) Through simulation benchmarking and real-world bimanual experiments on tasks such as laundry cleanup, fruit holding, and cluttered shelf picking, we demonstrate the superior performance of our method, particularly in bimanual manipulation tasks, compared to state-of-the-art diffusion-based approaches.\looseness=-1

\section{Related Works}

\noindent\textbf{Learning from Demonstrations:}
Learning from demonstrations is a growing area in robotics, spanning various methodologies and applications~\cite{ho2016generative, li2017infogail, hausman2017multi, fu2017learning}. Early research primarily focused on on-policy learning, where the agent interacts with and learns directly from the environment during training~\cite{pmlr-v15-ross11a}. To improve data efficiency, there has been a shift towards off-policy learning, employing strategies such as distribution matching~\cite{nachum2019dualdice, peters2010relative}, leveraging implicit network architectures~\cite{ibc}, and adopting more expressive networks~\cite{Chi2023DiffusionPV}. However, these methods often struggle with understanding and representing objects involved in manipulation tasks, crucial for accurate and effective action planning. We address this gap by adding state prediction as an additional supervision signal from human demonstrations to enhance data efficiency and improve object manipulation understanding.





\noindent\textbf{Model Learning in Robotics Manipulation:}
Dynamics models are crucial for complex robotic manipulation~\cite{bauza2017probabilistic, billard2019trends, zhou2019pushing}. Recent research primarily uses data-driven approaches to learn physical dynamics~\cite{chang2016compositional, nematollahi2020hindsight, baradel2020cophy, yin2021augmenting}. Promising results have been shown in modeling challenging dynamics such as deformable objects~\cite{yan2021learning, lin2022learning}, articulated objects~\cite{endres2013learning, mo2021where2act}, granular objects~\cite{wang2023dynamicresolution, chen2023learning}, and physical interactions under friction~\cite{chen2023predicting}. Particle scene representations and Graph Neural Networks are popular for modeling complex environments due to their adaptability to arbitrary geometries and ability to model physical interactions by aggregating particle interactions~\cite{mrowca2018flexible, li2019learning}. These methods often require expert design and are typically task-specific~\cite{wang2023dynamicresolution,chen2023predicting}. The learned dynamics models are used for state predictions in trajectory optimization within Model Predictive Control frameworks~\cite{shi2022robocraft, shi2023robocook}, or to replace costly environments in RL~\cite{ha2018world}. Our approach extends model learning to imitation learning by using human-collected datasets to simultaneously learn a state prediction model and an inverse dynamics model, enhancing generalization across tasks and improving versatility in robotic systems.

\noindent\textbf{Diffusion Models for Robotics:}
Diffusion models have been applied in robotics due to their expressiveness~\cite{Chi2023DiffusionPV, pearce2023imitating, janner2022diffuser}. They have been successfully applied in RL~\cite{ajay2023is, wang2023diffusion}, shared autonomy~\cite{yoneda2023diffusha}, imitation learning~\cite{Reuss2023GoalConditionedIL, xian2023chaineddiffuser}, motion planning~\cite{pmlr-v162-janner22a}, and reward learning~\cite{Huang2023DiffusionReward, nuti2023extracting}. The work most closely related to ours is Ajay et al.~\cite{ajay2023is}, who applied conditional diffusion models for state prediction and action generation in offline reinforcement learning (RL). {However, offline RL requires reward labeling and suffers from extrapolation error, which leads to the overestimation of out-of-distribution actions, causing instability during policy training~\cite{Fujimoto2018OffPolicyDR, kumar2019stabilizing}.}
Du et al.~\cite{du2023learning} demonstrated language-conditioned video generation through pretraining on large-scale text-video datasets with an inverse dynamics model for policy generation, but video generation often misses crucial low-level interaction information needed for precise manipulation, {such as contact dynamics, object deformation, and force application required for dexterous tasks.}
Moreover, their work focused solely on generating robot videos without conducting any robot experiments.
{Unlike previous works that focus primarily on simulation or video generation, our framework enables safe, reward-free deployment in real-world bimanual manipulation. Our method allows robots to solve complex, coordinated tasks while unlocking capabilities unattainable with simulation- or video-driven approaches.}\looseness=-1

\section{Approach}

\begin{figure}[tbp] 
  \centering
  \vspace{-1pt}
  \includegraphics[trim=0 0 40 0, clip, width=0.45\textwidth]{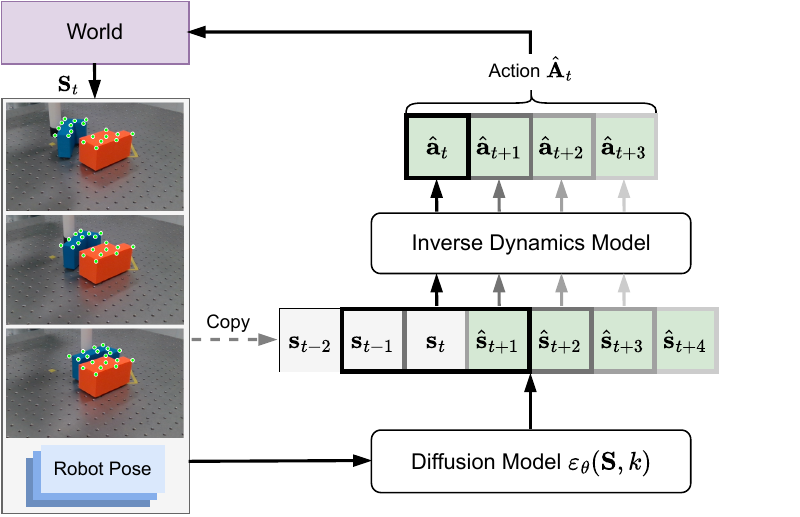}
  \caption{\textbf{Overview of the proposed framework.} At time step $t$, the Diffusion Model takes as input the latest $T_s$ steps of state data $\mathbf{S}_t$ and outputs the denoised future states. The resulting sequence of state is then sliced and processed by the inverse dynamics model to generate corresponding actions at each feasible time step within the prediction horizon. In the example of push-L task, the manipulated object state is characterized by a particle-based representation, as shown in the images on the left.  \vspace{-20pt}} 
  \label{fig:method}
\end{figure}

Our approach consists of two parts as shown in Figure~\ref{fig:method}: (1) A diffusion-based state prediction model to predict the future states of the world, (2) An inverse dynamics model that takes predictions as input to determine robot actions.

\noindent\textbf{State Prediction Diffusion Model:}
Our approach uses a variant of the Denoising Diffusion Probabilistic Model (DDPM)~\cite{ho2020denoising, nichol2021improved} to predict the state of the world at future timesteps. We use $\mathbf{S}_t$ to represent the state of the world at time $t$. Given an input with Gaussian noise $\mathbf{S}_t^k$, the goal is to recreate the true state at time \(t\), denoted as $\mathbf{S}_t^0$. The DDPM processes the noisy state \(\mathbf{S}^k_t\) along with the iteration index \(k\) to deduce the noise component, which is subtracted from the noisy state \(\mathbf{S}^k_t\). Through \(K\) iterations of denoising, the DDPM yields a sequence of intermediate states \(\mathbf{S}^k, \mathbf{S}^{k-1}, \ldots, \mathbf{S}^{0}\), each with progressively reduced noise, culminating in a noise-free state \(\mathbf{S}^0\). The reverse diffusion process can be represented as:
\begin{equation}
\mathbf{S}^{k-1}_t = \alpha(\mathbf{S}^k_t - \gamma\epsilon_\theta(\mathbf{S}^0_{t-T_s:t-1}, \mathbf{S}^k_t,k) + \mathcal{N} \bigl(0, \sigma^2 I \bigl))
\end{equation}
where \(\alpha\) and \(\gamma\) are derived from a closed-form expression of the variance schedule, and \(\epsilon_\theta(\mathbf{S}^0_{t-T_s:t-1}, \mathbf{S}^k_t, k)\) represents the model's estimate of the noise to be removed given the previous state history \(\mathbf{S}^0_{t-T_s:t-1}\). The Gaussian noise is denoted as \(\mathcal{N}(0, \sigma^2 \mathbf{I})\).

We train the diffusion-based state prediction model using the following loss function: 
\begin{equation}
\mathcal{L}_{pred} = \|\mathbf{\epsilon}^k - \epsilon_\theta(\mathbf{S}^0_{t-T_s:t-1}, \mathbf{S}^0_t + \mathbf{\epsilon}^k, k)\|_2^2
\end{equation}
where \(\mathcal{L}_{pred}\) is the \( L_2 \) loss between the actual noise \(\boldsymbol{\epsilon}^k\) and the noise predicted by the model \(\epsilon_\theta\), guiding the model in accurately estimating the noise to be removed at each iteration \(k\).

During training, states are randomly sampled from the demonstration dataset. For each sampled state, the network processes the denoising iteration \(k\), the noisy state, and preceding frames of the state to output the noise \(\epsilon_\theta\) that needs to be removed. This output is compared against a randomly sampled noise \(\epsilon\) with the appropriate variance for iteration \(k\), minimizing the loss as defined in Eq.~\ref{eq:diffusion_policy_loss}. In the deployment phase, we initialize a vector of length \(T_p\) with noise drawn from a Gaussian distribution, incorporating a low-temperature factor to reduce the initial noise level. After \(K\) iterations, the model generates predictions for future states, guiding the anticipated progression of the object or environment in relation to the task.

{The importance of modeling object movements through state prediction is highlighted in coordinated object manipulation. In such scenarios, if the object is dropped or mishandled, the state prediction loss would reflect this as a large state error. However, in state-to-action mapping, if the robot trajectories are similar but the object is dropped or not moved as expected, the action loss would remain relatively small, potentially missing the critical failure in object manipulation. Our method captures this distinction effectively, especially in tasks requiring bimanual coordination.}



\noindent\textbf{Inverse Dynamics Model:}
In the standard formulation of inverse dynamics models, inputs are states at two consecutive time steps, \(\mathbf{s}_t\) and \(\mathbf{s}_{t+1}\), and the output is the action required for the agent or robot to transition from state \(\mathbf{s}_t\) to \(\mathbf{s}_{t+1}\). We denote states \(\mathbf{s}_{t-T_h+1:t}\) as the history and \(\mathbf{s}_{t+1:t+T_f}\) as the future, respectively. We propose a modification to this model by introducing a variable number of historical and future states into the inverse dynamics framework. This modification allows the model to consider a sequence of past and future states, formalized as:
\begin{equation}
    \mathbf{a}_t = f^{-1} (\mathbf{s}_{t-T_h+1:t}, \mathbf{s}_{t+1:t+T_f})
\end{equation}
where \(T_h\) represents the number of historical states, and \(T_f\) denotes the number of future states. 

\noindent\textbf{Policy Composition:}
The state prediction network is adopted from the temporal convolutional neural network introduced by Janner et al.~\cite{janner2022diffuser} and Diffusion Policy~\cite{Chi2023DiffusionPV}. The inverse dynamics model is represented by a multi-layer perception model, which is trained using an MSE loss between the actual action and the model's predicted action $\mathcal{L}_{InvDyn}$. The final loss for our framework $\mathcal{L}$ is a composition of the state prediction loss and the action prediction loss, which optimizes both networks jointly. 
\begin{equation}
    \label{eq:diffusion_policy_loss}
    \begin{split}
    \mathcal{L}_{InvDyn}&=||{\mathbf{a}_t},f^{-1}(\mathbf{s}_{t-T_h+1:t+T_f} )||_2^2 \\
    \mathcal{L} &= {\beta} \cdot \mathcal{L}_{pred} + (1- {\beta}) \cdot \mathcal{L}_{InvDyn}
    \end{split}
\end{equation}
{We apply the inverse dynamics model on the input states and predicted states to generate \(T_a\) step actions, encouraging action consistency by incorporating temporal information into the decision-making process.}

  

\begin{figure}[tbp] 
  \vspace{0pt}
  \centering
  \includegraphics[width=0.4\textwidth]{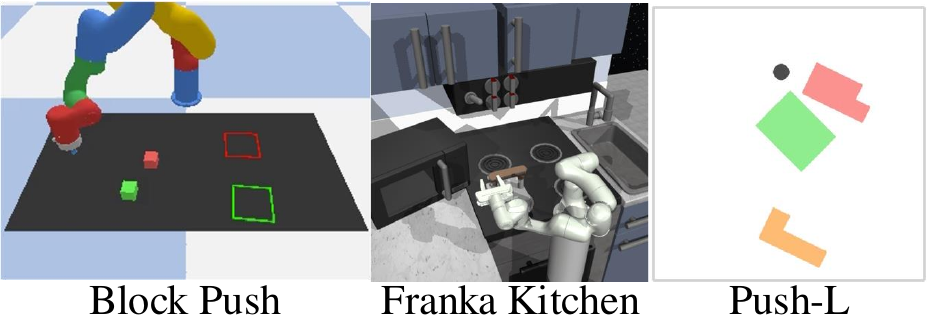}
\caption{\textbf{Simulation Benchmarks.} The XArm robot needs to push two blocks into randomized square positions. The Franka robot needs to manipulate seven objects in a virtual kitchen. The agent needs to push two L-shaped blocks to a target location. \vspace{-15pt}}
  \label{fig:simulation}
  \vspace{-5pt}
\end{figure}

\section{Simulation Benchmarking}


To validate our approach, we conducted extensive simulation experiments on various challenging tasks as part of a standard benchmarking process, as shown in Figure~\ref{fig:simulation}.
These experiments demonstrated the robustness and efficiency of our method against state-of-the-art baselines. We evaluated performance in diverse environments and datasets, focusing on tasks requiring precise manipulation and long-horizon planning. The results highlight our method's advantages in success rate, efficiency, and generalization across diverse scenarios. \looseness=-1


\subsection{Simulation Environments}

\begin{table*}[tb]
  \begin{center}
    \caption{\textbf{Performance of different models in both position (Pos) and velocity (Vel) control modes for the Franka Kitchen task.} Our policy shows superior performance across the task, interacting successfully with five objects and achieving a 29.3\% success rate in position control, despite training on a four-object interaction dataset.}
    \vspace{-5pt}
    \label{tab:franka_kitchen_results}
    \renewcommand{\arraystretch}{1.3}{
    \small
    \begin{tabular}{ l | c | c  c  c  c  c }
    \toprule
    \multirow{2}{*}{\textbf{Model}} & \multirow{2}{*}{\textbf{Ctrl}} & \multicolumn{5}{c}{\textbf{Franka Kitchen}} \\
                                    &                                & \textbf{p1}   & \textbf{p2}   & \textbf{p3}   & \textbf{p4}   & \textbf{p5}   \\
    \midrule
    DP     & Vel & 0.673$\pm$0.031 & 0.067$\pm$0.024 & 0.020$\pm$0.000 & 0.000$\pm$0.000 & 0.000$\pm$0.000 \\
    IDP   & Vel & 0.800$\pm$0.069 & 0.073$\pm$0.042 & 0.020$\pm$0.000 & 0.000$\pm$0.000 & 0.000$\pm$0.000 \\
    Ours                 & Vel & \textbf{0.920$\pm$0.020} & \textbf{0.427$\pm$0.083} & \textbf{0.133$\pm$0.031} & \textbf{0.027$\pm$0.012} & \textbf{0.007$\pm$0.012} \\
    \midrule
    DP     & Pos & \textbf{1.000$\pm$0.000} & \textbf{1.000$\pm$0.000} & \textbf{1.000$\pm$0.000} & \textbf{1.000$\pm$0.000} & 0.030$\pm$0.009 \\
    IDP   & Pos & \textbf{1.000$\pm$0.000} & \textbf{1.000$\pm$0.000} & \textbf{1.000$\pm$0.000} & 0.960$\pm$0.000 & 0.030$\pm$0.009 \\
    Ours                 & Pos & \textbf{1.000$\pm$0.000} & \textbf{1.000$\pm$0.000} & {0.993$\pm$0.012} & {0.953$\pm$0.042} & \textbf{0.293$\pm$0.012} \\
    \bottomrule
    \end{tabular}}
  \end{center}
  \vspace{-20pt}
\end{table*}

\noindent\textbf{Multimodal Block Pushing:} Adapted from Behavior Transformers~\cite{bet}, the Multimodal Block Pushing task requires an XArm robot to push two blocks into two squares in any order. Initial positions and rotations of the blocks are randomized.

    \begin{table}[tb]
      \caption{\textbf{Performance of different models in both position control (Pos) and velocity control (Vel) modes for the Block Push task.} Our model achieves higher p1 and p2 metrics in both control modes compared to other diffusion models.}
      \vspace{-10pt}
      \begin{center}{
        \begin{tabular}{ l | c | c  c }
        \toprule
        \multirow{2}{*}{\textbf{Model}} & \multirow{2}{*}{\textbf{Ctrl}} & \multicolumn{2}{c}{\textbf{Block Push}} \\
                                        &                                & \textbf{p1}    & \textbf{p2}    \\
        \midrule
        DP     & Vel & 0.353$\pm$0.034 & 0.107$\pm$0.025 \\
        IDP   & Vel & 0.327$\pm$0.025 & 0.120$\pm$0.016 \\
        Ours                 & Vel & \textbf{0.633$\pm$0.068} & \textbf{0.280$\pm$0.016} \\
        \midrule
        DP     & Pos & 0.648$\pm$0.031 & 0.247$\pm$0.012 \\
        IDP   & Pos & 0.647$\pm$0.081 & 0.327$\pm$0.063 \\
        Ours                 & Pos & \textbf{0.713$\pm$0.012} & \textbf{0.340$\pm$0.060} \\
        \bottomrule
        \end{tabular}}
      \end{center}
      \label{tab:block_push_results}
      \vspace{-15pt}
    \end{table}

\noindent\textbf{Franka Kitchen:} Adapted from \cite{gupta2019relay}, the Franka Kitchen task requires a Franka robot to manipulate 7 objects (a microwave, kettle, slide cabinet, hinge cabinet, light switch, and two burner knobs) in a virtual kitchen environment.


\noindent\textbf{Push-L:}
Adapted from Implicit Behavior Cloning and diffusion policy~\cite{ibc, Chi2023DiffusionPV}, the Push-L task requires manipulating two L-shaped blocks toward a target using a circular end-effector. The task is inherently long-horizon and multi-stage, requiring the strategic assembly of the two objects before they navigate to the goal area. The agent must exploit complex, contact-rich multi-object dynamics for precise manipulation. The symmetric shape of the combined objects allows for multiple goal configurations to achieve success. Variability is introduced through randomized initial positions of the blocks and end-effector.

    \begin{table}[tbp]
      \caption{\textbf{Comparison of models on the Push-L task.} Success rates demonstrate that our model with the inverse dynamics model significantly outperforms baselines.}
      \vspace{-5pt}
      \begin{center}{
        \begin{tabular}{ l | c  }
        \toprule
        \textbf{Model} & \textbf{Push-L} \\
        \midrule
        RNN-GMM &  0.120 ± 0.035 \\
        DP & 0.474 ± 0.025 \\
        IDP & 0.502 ± 0.012 \\
        Ours w/o Inv & 0.613 ± 0.034 \\
        Ours & \textbf{0.793 ± 0.042} \\
        \bottomrule
        \end{tabular}}
      \end{center}
      \label{tab:push_l_results}
      \vspace{-25pt}
    \end{table}


\subsection{Baselines and Evaluation Metrics}
We employ the DP (Diffusion Policy), which utilizes a diffusion model for end-to-end state-to-action mapping~\cite{Chi2023DiffusionPV}. Additionally, we integrate recent advancements~\cite{nichol2021improved, Ning2023InputPR} in diffusion model research to enhance the performance of the DP, resulting in what we refer to as the IDP (Improved Diffusion Policy). The modification aims to benchmark the foundational capabilities and improvements over the standard model, providing a comparative analysis of the performance enhancements from our framework. {To demonstrate the necessity of the denoising network, we also include a baseline RNN-GMM (Recurrent Gaussian Mixture Model) policy.}
We report the results from the best-performing checkpoint across {50 different initial conditions from 3 seeds (150 in total)}, with the metric being the success rate for all tasks. Moreover, we examine each method's effectiveness in both position control (Pos) and velocity control (Vel) modes.

\textbf{Evaluation Metrics:} {We use the success rate to measure how effectively the agent pushes two L-shaped blocks to the target areas in the Push-L task. The p'' values in the Block Push Environment represent the number of blocks pushed into the target region. In the Franka Kitchen Task, the p'' values indicate the number of tasks successfully completed, serving as a quantitative measure of task accomplishment. 
These tasks include turning the oven knobs, turning on the light switch, opening the slide cabinet, opening the hinge cabinet, opening the microwave door, and moving the kettle to the top left burner.}


\subsection{Results}

\noindent\textbf{Capability for Task Behavior Synthesis:}
The performance outcomes for the Franka Kitchen tasks are detailed in Table~\ref{tab:franka_kitchen_results}. Our approach surpasses the baseline models across all control configurations. {Notably, in 29.3\% of the trials, our position control policy successfully completed 5 tasks. Although the dataset contains trajectories where the robot achieves 4 out of these 7 tasks in a random order, our policy is able to synthesize the behavior to achieve 5 tasks out of 7.}
Such results underscore our model's advanced capability to synthesize complex behaviors that extend beyond the scope of the initial demonstrations, highlighting its potential for adaptive and intelligent robotic control in multi-stage multi-task environments.

\noindent\textbf{Necessity of an Explicit Inverse Dynamics Model:}
We evaluate the importance of an explicit inverse dynamics model by examining its impact on performance in the Push-L task, as shown in Table~\ref{tab:push_l_results}. Our approach achieves a 79.3\% success rate, while removing the inverse dynamics model resulted in an 18\% performance drop. Although the state prediction model can output the future state sequence, including the pusher's position, it fails to capture the nuances of how the agent interacts with objects. Integrating the inverse dynamics model allows the robot to better understand the effect of contact dynamics on state transitions, leading to a more comprehensive insight into agent-object interactions.

\noindent\textbf{Impact of Control Mode on Policy Performance:}
The performance outcomes for the Multimodal Block Pushing task are detailed in Table~\ref{tab:block_push_results}. The benchmark results compare the performance across different models. Our method significantly outperforms both the original and improved diffusion policies, showing a substantial increase in the number of objects successfully manipulated (p1 and p2 metrics). Notably, our approach maintains stability between velocity and position control modes, avoiding the performance degradation seen in the end-to-end diffusion policy.

\begin{figure}[tbp] 
    \centering %
    \caption{\textbf{Performance comparison in simulation between ours and improved diffusion policy.} The x-axis represents the dataset size, and the y-axis represents the success rate. Our method demonstrates higher sample efficiency as it achieves better performance with the same dataset size due to the utilization of more supervision signals.}\vspace{-7pt}
    \label{fig:data_efficiency}
    \includegraphics[trim=0 0 30 0, clip, width=0.8\linewidth]{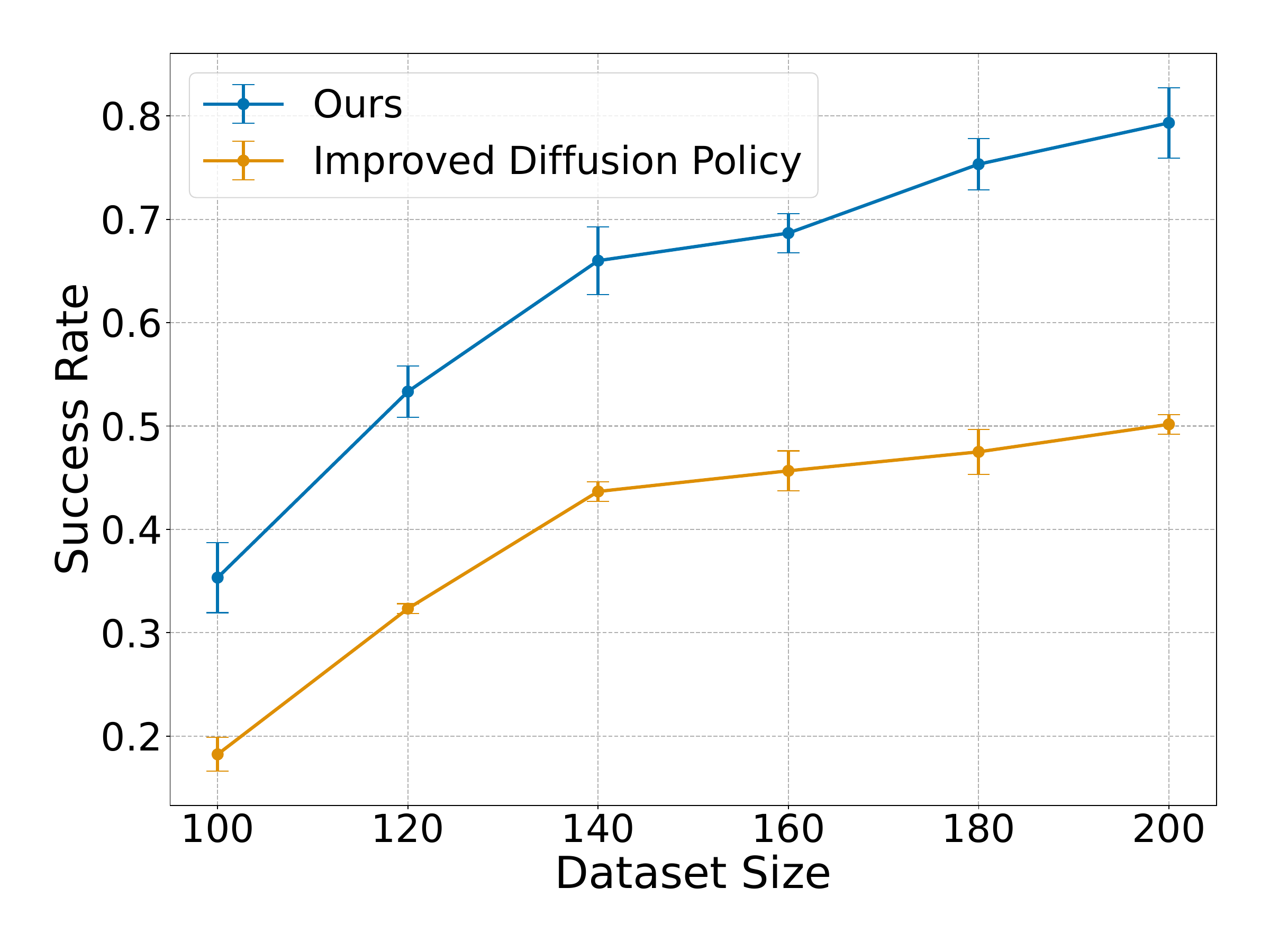}\vspace{-30pt}
\end{figure}
\noindent\textbf{Data Efficiency:}
We analyze the data efficiency of our proposed method compared to the Improved Diffusion Policy in simulation, and the results are shown in Figure~\ref{fig:data_efficiency}.
Our method consistently outperforms the Improved Diffusion Policy across all dataset sizes. For example, with a dataset size of 100 {demonstration trajectories}, our method achieves a success rate of 0.32, compared to 0.17 for the Improved Diffusion Policy. At a dataset size of 200, our method's success rate increases to 0.793, while the Improved Diffusion Policy reaches 0.502. The superior data efficiency of our method stems from incorporating more supervision signals during training. These additional signals provide comprehensive information, enabling the model to learn more effectively from each data point. Consequently, our method generates more effective action plans with a smaller dataset, leading to higher success rates in manipulation tasks. This efficiency is particularly beneficial in real-world applications, where data collection can be resource-intensive and expensive.

\section{Real-World Bimanual Coordination Experiments}
\label{sec:result}

We evaluate our model's real-world performance on bimanual coordination tasks, demonstrating its ability to tackle long-horizon tasks, generate smooth state trajectories, and manage action discontinuities. We performed a zero-shot sim-to-real transfer of the push-L task, training the policy on a simulation dataset. Additionally, we trained the policy on real-world tasks: laundry cleanup, fruit holding, and cluttered shelf picking. \looseness=-1

\subsection{2D Push-L Experiment}

\begin{figure*}[t]
  \vspace{-10pt}
  \centering
  \includegraphics[width=0.96\linewidth]{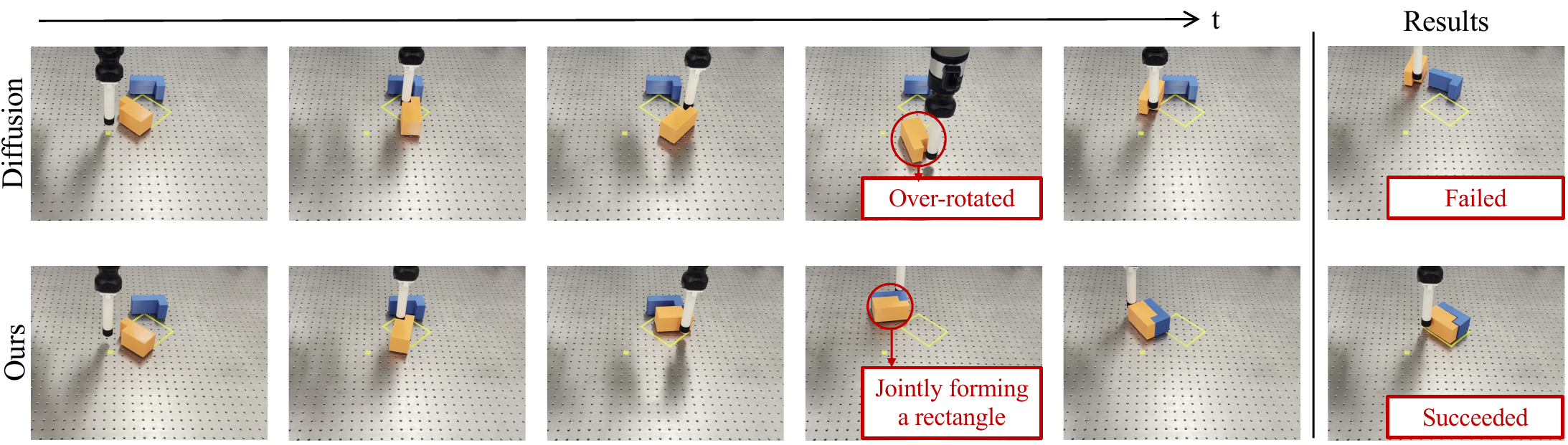}
  \caption{\textbf{Real-world comparison of different models on the Push-L task.} In the first row, the improved diffusion model at first pushes the orange block towards the target position, but then over-rotates this block, resulting in a failed trial. In the second row, our model pushes the orange block towards the blue block to form a joint rectangle. The agent then pushes the joint rectangle toward the target location successfully. \vspace{-7pt}}
  \label{fig:quali}
\end{figure*}

\noindent\textbf{Real-World Push-L Task:}
We evaluate our method in the real world using policy trained from the simulation demonstration dataset. Initially, we extract the key points of objects based on their poses within the real-world coordinate system, utilizing our perception pipeline. These key points, identifiable on the blocks, can be visualized as in Figure~\ref{fig:method}. Subsequently, we convert these key points into the simulation coordinate system, serving as inputs to our model. The model predicts the corresponding future states and actions within the simulation. To actuate the robot, we translate the simulation-derived actions back to the real-world context through an inverse transformation process, mapping them onto the robot base coordinate system. This procedure establishes a bidirectional link between the real-world setup and the simulation, enabling seamless task execution.

\noindent\textbf{Quantitative Results:}
We conducted tests under 12 distinct initial conditions, as depicted in Figure~\ref{fig:pushl_initial_setup}. These conditions were designed to cover a comprehensive range of scenarios, including varying distances between the blocks and their target regions, differences in the proximity of the blocks to one another, assorted rotational configurations, and diverse relative poses between the blocks and the agent. Our method achieves 9 out of 12 successful outcomes, significantly outperforming the Improved Diffusion model, which succeeded in only 2 of the 12 trials. This result indicates our model's capability to synthesize reasonable plans for object movement, consequently generating effective actions to manipulate the objects toward the desired outcomes.

\noindent\textbf{Qualitative Results:}
In our real-world demonstration of the Push-L task, our policy predicted the {future action sequence} the agent should take. In Figure~\ref{fig:quali}, from the initial position of the blocks, the agent began by first pushing the orange block towards the blue block. The agent then adjusted the position of the orange block so it could form a joint rectangle with the blue block. This newly formed joint rectangle increases the ease of manipulation and can then be pushed to the target location efficiently. We compare these findings with Diffusion Policy. With Diffusion Policy, the agent pushes the orange block towards the target location but then over-rotates the orange block. This causes the trial to fault and the two blocks are not able to reach the target location.

\subsection{3D Bimanual Experiments}
We additionally test our method in a challenging real-world bimanual manipulation tasks.
We collected 200 demonstrations on UR5e robots using the Gello~\cite{Wu2023GELLOAG} teleoperation system for each task.


\noindent\textbf{Laundry Cleanup:} 
In the Laundry Cleanup task, we removed the grippers from our robots and the robots need to move two pillows from the table to the laundry basket. This task is challenging due to the need for precise coordination and handling of soft, deformable objects. The observation is point cloud of the robot and the scene. The actions involve controlling the joint positions of both robots. 

\noindent\textbf{Fruit Holding:} 
In the Fruit Holding task, four fruits are placed on a table, and the robots must hold two fruits at a time and move them to a bowl. This simulates the challenge of humans holding multiple fruits, especially when they are larger than the gripper openings, testing the robots' ability to handle objects of varying sizes. The observation is the point cloud of the robot and scene, and the actions involve controlling the joint positions of both robots.

\noindent\textbf{Cluttered Shelf Picking:} 
In the Cluttered Shelf Picking task, a shelf is filled with various objects. 
One robot needs to pick a target object from the shelf, while the other robot prevents non-target objects from moving or collapsing. This task is particularly challenging due to the need for spatial awareness in a densely packed environment. The observation is point cloud of the robot and the scene. The actions involve controlling the joint positions and grippers of both robots. 

\subsection{Results}

\begin{table}[t]
  \begin{center}
      \caption{Performance comparison between our method and improved diffusion. LC: Laundry Cleanup (p\# = number of pillows moved). FH: Fruit Holding (p\# = number of fruits moved). CSP: Cluttered Shelf Picking (successful trials).\vspace{0pt}}
    \label{tab:real_world_results}
    \begin{tabular}{l|cc|cccc|c}
    \toprule
    \multirow{2}{*}{} & \multicolumn{2}{c|}{\textbf{LC}} & \multicolumn{4}{c|}{\textbf{FH}} & \textbf{CSP} \\
                      & p1   & p2   & p1   & p2   & p3   & p4   & Success \\
    \midrule
    IDP & 14/15 & 0/15 & 14/15 & 14/15 & 10/15 & 5/15 & 0/15 \\
    Ours               & \textbf{15/15} & \textbf{8/15} & \textbf{15/15} & \textbf{15/15} & \textbf{15/15} & \textbf{8/15} & \textbf{14/15} \\
    \bottomrule
    \end{tabular}
  \end{center}
  \vspace{-25pt}
\end{table}

\begin{figure*}[tbp]
  \centering
  \includegraphics[width=.92\linewidth]{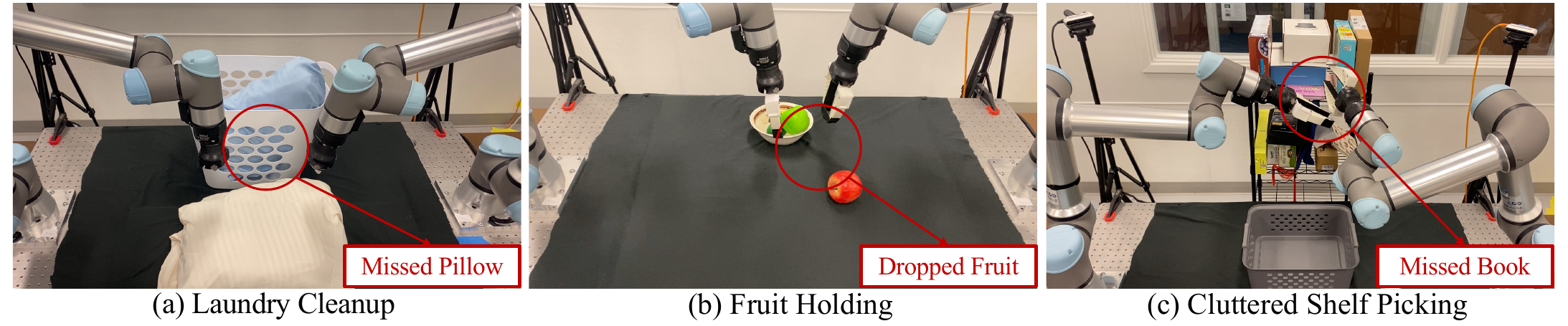}
  \caption{\textbf{Typical failure cases of the baselines.} In the laundry cleanup task, the action-mimicking diffusion baseline failed to capture object movements and dropped the second pillow; in the fruit holding task, the baseline dropped the fruit during transportation; in the cluttered shelf task, the baseline missed extracting the book from the shelf.\vspace{-5pt}}
  \vspace{-10pt}
  \label{fig:initial_states}
\end{figure*}


Figure~\ref{fig:initial_states} illustrates the typical failure cases of the baselines, showing their limitations in coordinated bimanual tasks. In the laundry cleanup task, the action-mimicking diffusion baseline failed to accurately capture object movements, resulting in the second pillow being dropped. During the fruit holding task, the baseline struggled to maintain a grip on the fruit during transportation, leading to it being dropped. In the cluttered shelf task, the baseline was unable to extract the book from the shelf, demonstrating its difficulty in handling complex object interactions that require bimanual coordination.
Table~\ref{tab:real_world_results} presents a detailed performance comparison between our method and the Improved Diffusion model. We conducted each test under 15 different initial conditions, covering a comprehensive range of scenarios, including multi-object handling, modeling interactions between the robot and objects, whole-body control, and deformable objects using bimanual robots.
In the Laundry Cleanup task, our method achieved a perfect score of 15/15 for moving the first pillow (p1) and 8/15 for moving the second pillow (p2), whereas the Improved Diffusion model achieved 14/15 for p1 and 0/15 for p2. This demonstrates that our method is highly effective in whole-body control and modeling the interactions between the robot's links and soft objects like pillows. The coordination between the two robots was also successful.
For the Fruit Holding task, our method outperformed the Improved Diffusion model across all scenarios (p1, p2, p3, and p4). Our method achieved a perfect score of 15/15 for p1, p2, and p3, and 8/15 for p4. In contrast, the Improved Diffusion model achieved 14/15 for p1 and p2, 10/15 for p3, and only 5/15 for p4. These results underscore our method's robustness in handling complex manipulation tasks involving multiple objects and varying configurations. The higher success rates demonstrate our method's ability to effectively model multi-object interactions, plan accurately, and execute precise movements, leading to successful task completion.
For the Cluttered Shelf Picking task, our method showed superior performance, as it better models interactions between the gripper and objects compared to the baseline. 


\section{Conclusions and Future Work}
In this paper, we present a novel approach for coordinated bimanual manipulation utilizing a state prediction diffusion model, which takes observation sequences and predicts future states. Further, an inverse dynamics model is employed to translate these predicted observations into specific actions for the agent. Our evaluations across diverse complex environments, including Block Push, Franka Kitchen, and Push-L tasks, demonstrate superior performance against other state-of-the-art end-to-end policies. Notably, in the Push-L task, our approach achieves a higher success rate, showing remarkable adaptability to varying initial block positions. We further demonstrate the approach on real-world challenging bimanual coordinated tasks, including laundry cleanup, fruit holding, and cluttered shelf picking. \looseness=-1
The results underscore our method’s robust capability in coordinating dual arms for stable and precise control in bimanual manipulation, where understanding scene evolution is crucial. Future research could focus on optimizing the training process to reduce time requirements, improving data handling efficiency, and streamlining the collection of real-world demonstrations to enhance practicality and scalability in coordinated bimanual tasks.

\noindent\textbf{Limitations: } 
Our work encompasses the following limitations. 
First, our method takes longer time to train than the diffusion policy~\cite{Chi2023DiffusionPV} because the state space is typically larger than the action space, making the training longer. Second, collecting real-world demonstrations is time-consuming and labor-intensive.\looseness=-1




\section*{Acknowledgments}
{We thank Yiyang Xu's assistance with the demonstration collection, as well as Zhe Huang, and Neeloy Chakraborty for his insightful feedback and suggestions.
This work was supported by ZJU-UIUC Joint Research Center Project  No. DREMES 202003, funded by Zhejiang University.
This work utilizes resources supported by the National Science Foundation’s Major Research Instrumentation program, grant \#1725729, as well as the University of Illinois at Urbana-Champaign.
This research used the Delta advanced computing and data resource which is supported by the National Science Foundation (award OAC 2005572) and the State of Illinois. Delta is a joint effort of the University of Illinois Urbana-Champaign and its National Center for Supercomputing Applications.
Additionally, this work used the Delta system at the National Center for Supercomputing Applications through allocation ELE230010 from the Advanced Cyberinfrastructure Coordination Ecosystem: Services \& Support (ACCESS) program, which is supported by National Science Foundation grants \#2138259, \#2138286, \#2138307, \#2137603, and \#2138296.
}

\bibliographystyle{IEEEtran}
\bibliography{BibFile}
\clearpage

\newpage
\section*{\Large Appendices} 


\label{sec:appendices}
\renewcommand{\thesubsection}{\Alph{subsection}}


\subsection{Optimizing Over the Whole Trajectory}

In behavior cloning, we assume we have expert or rational demonstrations. The network's goal is to generate the corresponding action given a state. During training, we maximize the probability of the action from the dataset, assuming that each step's action is optimal.

The objective function for behavior cloning can be formulated as follows: Given a dataset of state-action pairs \(\{(s_t, a_t)\}\), we define the probability \(p(a_t \mid s_t)\) of taking action \(a_t\) given state \(s_t\).
The goal is to maximize the log-probability of the actions over all state-action pairs in the dataset:

\[
\max_{\theta} \frac{1}{N} \sum_{t=1}^{N} \log p_\theta(a_t \mid s_t)
\]

Here, \(p_\theta(a_t \mid s_t)\) represents the probability of taking action \(a_t\) given the state \(s_t\) and the model parameters \(\theta\). By optimizing \(\theta\), we aim to increase the likelihood of the actions in the dataset, thereby learning a policy that can imitate the behavior demonstrated in the dataset.

However, behavior cloning has a significant limitation: it fails to consider the entire trajectory, potentially getting stuck in locally-optimal or suboptimal states. 

To address its limitation, we assume human-collected trajectories represent either the lowest cost or highest reward scenarios. Given a sequence of states \( s_1, s_2, \ldots, s_T \) and actions \( a_1, a_2, \ldots, a_{T-1} \), the joint probability can be factorized sequentially as follows:
\begin{align*}
p(s_1, a_1, s_2, a_2, \ldots, s_T) = p(s_1) \prod_{t=1}^{T-1} p(s_{t+1} \mid s_t) \nonumber \\
\cdot p(a_t \mid s_{t-T_h+1:t}, s_{t+1:t+T_f})
\end{align*}
Taking the logarithm of both sides, we obtain:
\begin{align*}
&\log p(s_1, a_1, s_2, a_2, \ldots, s_T) = \log p(s_1) \nonumber \\
&+ \sum_{t=1}^{T-1} \left( \log p(s_{t+1} \mid s_t) + \log p(a_t \mid s_{t-T_h+1:t}, s_{t+1:t+T_f}) \right)
\end{align*}

We make the following assumptions:
\begin{itemize}
    \item \( p(s_1) \): The probability of the initial state.
    \item \( p(s_{t+1} \mid s_t) \): The effect model, representing state transitions.
    \item \( p(a_t \mid s_{t-T_h+1:t}, s_{t+1:t+T_f}) \): The inverse dynamics model, representing the dependencies on previous and future states.
\end{itemize}

Assuming the human-collected trajectory is optimal, our goal is to maximize the probability of the entire trajectory. By applying the logarithm to the probability, we derive the corresponding objective functions.

To maximize the log probability of the entire trajectory, we combine the objectives as follows:

\begin{align*}
\max \Bigg( \log p(s_1) + \sum_{t=1}^{T-1} \log p(s_{t+1} \mid s_t) \nonumber \\
+ \sum_{t=1}^{T-1} \log p(a_t \mid s_{t-T_h+1:t}, s_{t+1:t+T_f}) \Bigg)
\end{align*}

Since the probability of the initial state \( p(s_1) \) is often known or assumed to be fixed, we focus on optimizing the state transitions and actions:
\begin{align*}
\max \Bigg( \sum_{t=1}^{T-1} \log p(s_{t+1} \mid s_t) &+ \\
\sum_{t=1}^{T-1} \log p(a_t \mid s_{t-T_h+1:t}, s_{t+1:t+T_f}) \Bigg)
\end{align*}
This final objective function captures the essence of optimizing the trajectory based on state transitions and actions, assuming the initial state probability is given.

\subsection{Implementation Details}
To obtain a low-dimensional state representation in the real world, we first train an encoder guided by an inverse dynamics model to compress the high-dimensional input. We then train the state prediction diffusion model and fine-tune the inverse dynamics model. 

\begin{table}[tb]
  \begin{center}
      \caption{\textbf{Summary of Hyperparameters for Different Tasks.} $T_p$: state vector length, $T_s$ : True state length, $T_a$ : Action execution horizon, $T_h$ : History input length toward the inverse dynamic model, $T_f$ : Future input length toward the inverse dynamic model.}
    \label{tab:hyperparameters}
    \begin{tabular}{ l | c c c c c c }
    \toprule
    Task & Ctrl & \textbf{$T_s$} & \textbf{$T_a$} & \textbf{$T_p$} & \textbf{$T_h$} & \textbf{$T_f$} \\
    \midrule
    Push-L          & Pos & 2 & 4 & 16 & 1 & 2 \\
    Block Push      & Pos & 3 & 1 & 12 & 1 & 2 \\
    Franka Kitchen  & Pos & 2 & 4 & 16 & 2 & 1 \\
    Block Push      & Vel & 3 & 2 & 12 & 1 & 2 \\
    Franka Kitchen  & Vel & 2 & 4 & 16 & 1 & 2 \\
    \bottomrule
    \end{tabular}
  \end{center}
  \vspace{-10pt}
\end{table}

\noindent\textbf{Hyper-Parameter Selection:}
Hyperparameter Tuning is crucial for our model's performance across a variety of tasks. In this section, we outline the hyperparameter configurations tailored for each task, as summarized in Table~\ref{tab:hyperparameters}. For tasks with human demonstrations, like Push-L and Franka Kitchen, we set the state vector length $T_p$ to 16, the true state length $T_s$ to 2, and the action execution horizon $T_a$ to 4. In contrast, the Block Push Task, derived from Oracle script demonstrations, requires a distinct setup: an state vector length $T_p$ of 12, a true state length $T_s$ of 3, and an action execution horizon $T_a$ of 1 for position control, with $T_a=2$ for velocity control. When considering the input length to the inverse dynamics model, a history length $T_h$ of 1 and a future length $T_f$ of 2 generally result in an optimal performance. However, the Franka Kitchen task under position control benefits from a longer historical context $T_h=2$, indicating the need for more extensive historical data to synthesize more extended behavior sequences effectively.

\begin{figure*}[tb]
  \centering
  \includegraphics[width=\linewidth]{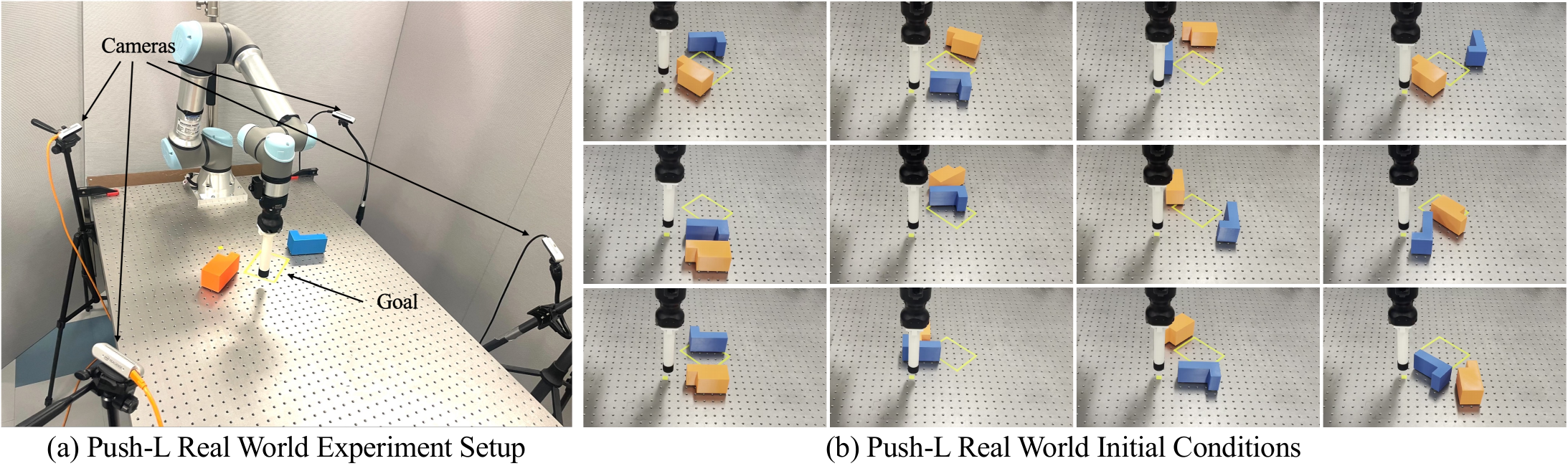}
  \caption{(a) {\textbf{Real-world experiment setup.} We position four cameras at all four corners to fully capture the workspace. The goal location is marked on the table with yellow tape.} (b) \textbf{Different initial conditions for the real-world Push-L task.} To assess the efficacy of our policy compared to the improved diffusion policy, we established 12 distinct block configurations. These setups were designed to span a broad spectrum of scenarios, encompassing variations in block-to-target distances, block proximities, rotational states, and the relative positioning of blocks to the agent. \vspace{-0pt}}
  \vspace{-5pt}
  \label{fig:pushl_initial_setup}
\end{figure*}

\noindent\textbf{Real-World Robot Setup of Push-L Task:}
{Our experimental setup includes a UR5e robot and 4 RealSense D415 cameras as shown in Figure~\ref{fig:pushl_initial_setup} (a). We position the 4 cameras to be angled and facing towards the workspace area to completely capture the blocks' movements and the block locations. We observe frame rate discrepancy for sim-to-real transfer, with the real-world operations proceeding at a frame rate of 1, in contrast to the simulation's frame rate of 10. Due to the lagging from the perception system and control latency, the interval between consecutive frames in the real world was observed to be narrower than that in the simulation.} 

\begin{figure*}[tb]
  \centering
  \includegraphics[width=\linewidth]{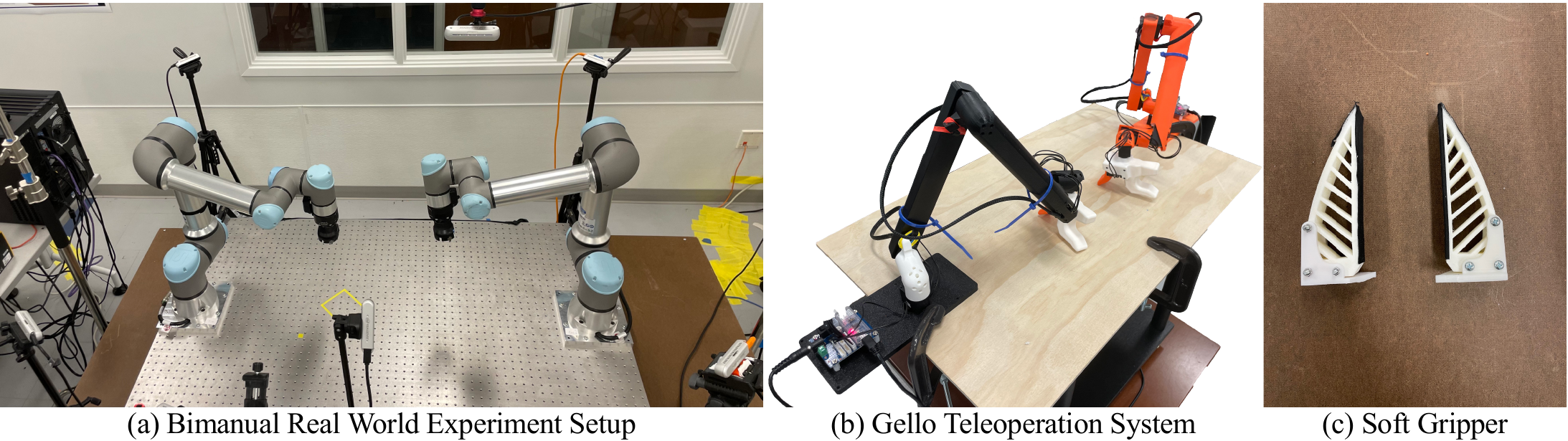}
  \caption{\textbf{Bimanual Robot Setup.} (a) Bimanual experimental setup with two UR5e robots. (b) Gello teleoperation system. (c) The soft gripper model for the Robotiq Hand-E gripper, made from TPU material and featuring anti-slip tape for increased friction.}
  \vspace{-5pt}
  \label{fig:bimanual_setup}
\end{figure*}

\noindent\textbf{Real-World Perception Module for Push-L Task:}
After collecting RGB and depth images from the 4 calibrated cameras, we convert them to colored point clouds. This point cloud is aggregated in the world coordinate system, after which we crop it to focus on the workspace area. Subsequent steps involve employing color segmentation techniques to isolate points corresponding to the object and utilizing voxel downsampling to reduce the point cloud's density. We proceed to eliminate any outliers and compute the normals of the point cloud. The processed point cloud allows us to perform surface matching and the Iterative Closest Point (ICP) registration~\cite{5540108}. We find the transformation to align between the point cloud of the blocks, captured in real-time, and the surface point cloud sampled from the object mesh, thereby calculating the object pose in real time.

\noindent\textbf{Bimanual Robot Setup:}
Our experimental setup features a bimanual configuration with two UR5e robots, illustrated in Figure~\ref{fig:bimanual_setup} (a). The Gello teleoperation system, referenced from Wu et al.~\cite{Wu2023GELLOAG}, is shown in Figure~\ref{fig:bimanual_setup} (b). This system captures joint values from Gello and transmits them to the UR5e robots during teleoperation. The soft gripper, designed based on the model from Chi et al.~\cite{Chi2024UniversalMI}, is used with the Robotiq Hand-E gripper and is depicted in Figure~\ref{fig:bimanual_setup} (c). The gripper is constructed from TPU material and is equipped with anti-slip tape to increase friction.

\begin{figure}[tb]
  \centering
  \includegraphics[width=\linewidth]{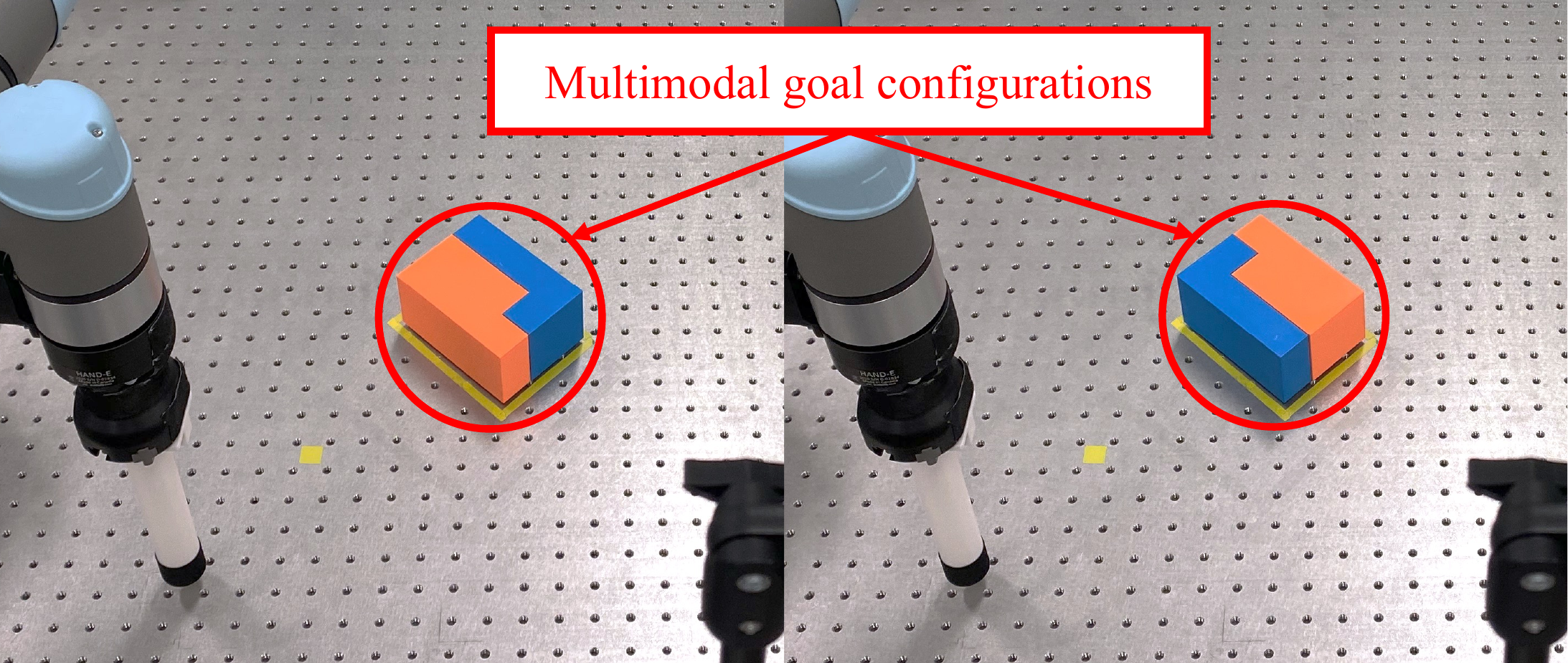}
    \caption{\textbf{Multimodal goal configurations.} This figure depicts the goal configurations the agent must solve in the Push-L task. The multiple goal configurations allow for several possible object and robot action paths that the agent can perform to achieve success.}
  \label{fig:multi_goal}
\end{figure}

\subsection{Additional Results}

\subsubsection{Simulation Dataset}
\begin{itemize}
    \item \noindent\textbf{Multimodal Block Pushing:}
    A total of 1,000 demonstrations were generated using a scripted oracle. In the demonstrations, the agent reaches red and green blocks and pushes them towards either square with equal probability. Observations include positions and rotations of the two blocks, current end-effector position, targeted end-effector position, and goal regions' positions and rotations. Actions can be the effector's future position or the desired relative movement.
    \item \noindent\textbf{Franka Kitchen:}
    The dataset contains 566 demonstrations by human experts. Each demonstration includes the completion of 4 tasks in an arbitrary order, and the robot is expected to achieve as many demonstrated tasks as possible. Observations include the object state and robot joint state, while actions are represented by the robot's joint angles or joint velocities.
    \item \noindent\textbf{Push-L:}
    We collect a dataset of 200 demonstrations. Observations include key points from the two L-shaped blocks and the agent's position. Actions are defined by the agent's future position.
\end{itemize}

\subsubsection{Push-L Manipulation Experiment}

\begin{figure*}[tb]
  \centering
  \includegraphics[width=\linewidth]{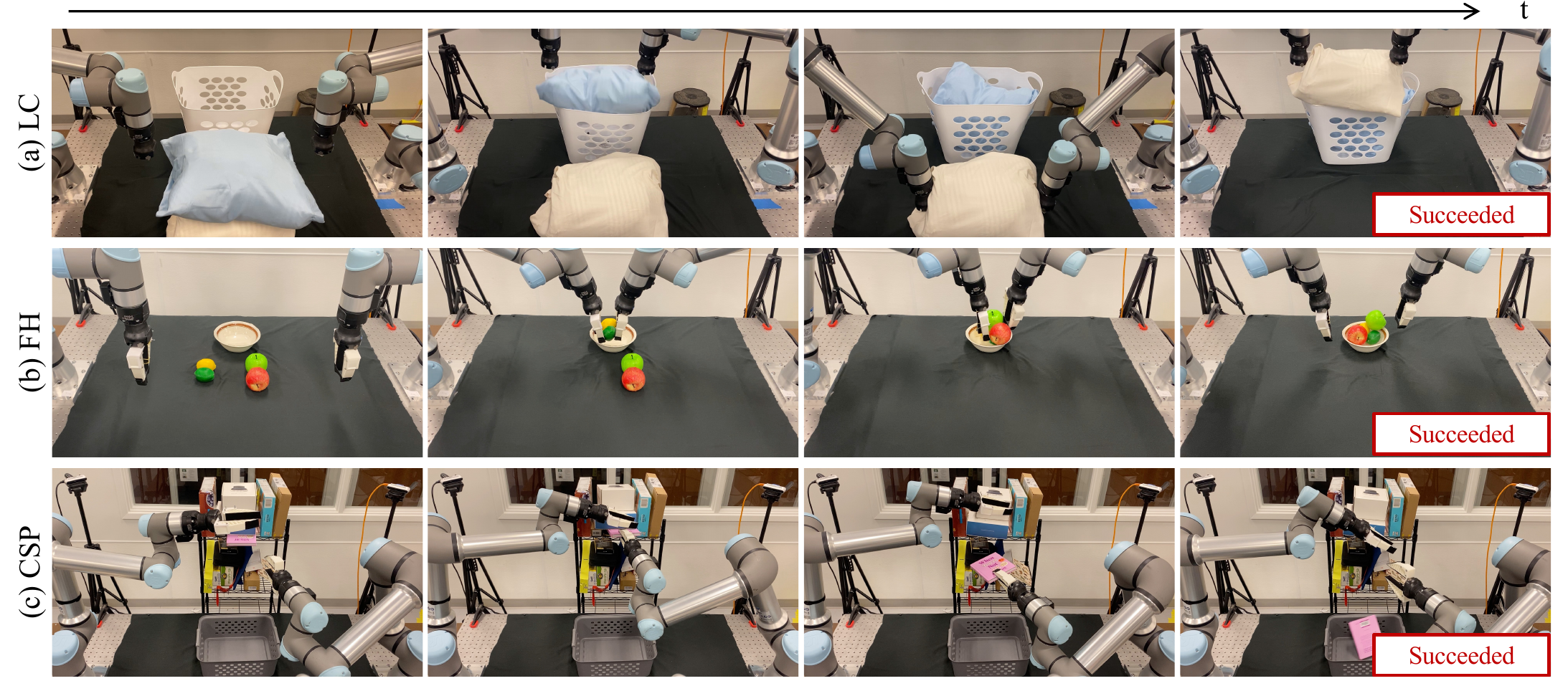}
  \caption{\textbf{Qualitative results from our real-world experiments.} (a) During the laundry cleanup task, our state-predictive method allowed the robot to handle the second pillow effectively. (b) In the fruit holding task, our method moved all four fruits without dropping any. (c) For the cluttered shelf picking task, our method retrieved the pink book from the shelf without disturbing other items. These outcomes underscore the robustness and versatility of our approach in practical scenarios.}
  \label{fig:quali_real}
\end{figure*}

\begin{figure*}[tb]
  \centering
  \includegraphics[width=\linewidth]{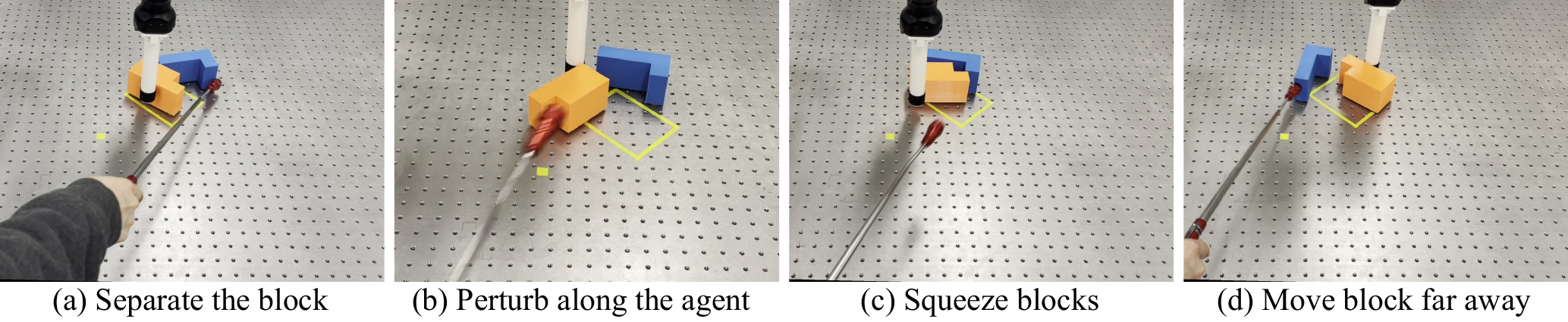}
  \caption{\textbf{Human-performed perturbation scenarios.} (a) Separate the block: a human moved one block away from its original position. (b) Perturb along the agent: a human adjusted the block along the path of the agent. (c) Squeeze the block: a human squeezed the block, altering the path of both blocks. (d) Move block far away: a human relocated the block to a distant position.}
  \label{fig:pushL_human}
  \vspace{-5pt}
\end{figure*}

\begin{itemize}
    \item \noindent\textbf{Characteristics and Complexity of Push-L Task:}
    The Push-L task is contact-rich, involving intricate interactions between objects and between the objects and the agent. An agent is tasked with maneuvering two L-shaped objects to a target location. A key aspect of this environment is the presence of multiple feasible action trajectories, each capable of successfully completing the task. This complexity is further enhanced by the requirement for the agent to achieve success through a variety of goal configurations, as shown in Figure~\ref{fig:multi_goal}.
    \item \noindent\textbf{Robustness Testing:}
    For robustness testing, a human performed various perturbations on the system. In one scenario, the human moved the block away from its original position. Another test involved the human adjusting the block along the agent's path. Additionally, the human compressed the block, altering its path, and finally, relocated the block to a significantly distant position. These tests are illustrated in Figure~\ref{fig:pushL_human}. Our method successfully handled all these perturbations, demonstrating its robustness and adaptability.

\end{itemize}

\begin{figure*}[tb]
  \centering
  \includegraphics[width=\linewidth]{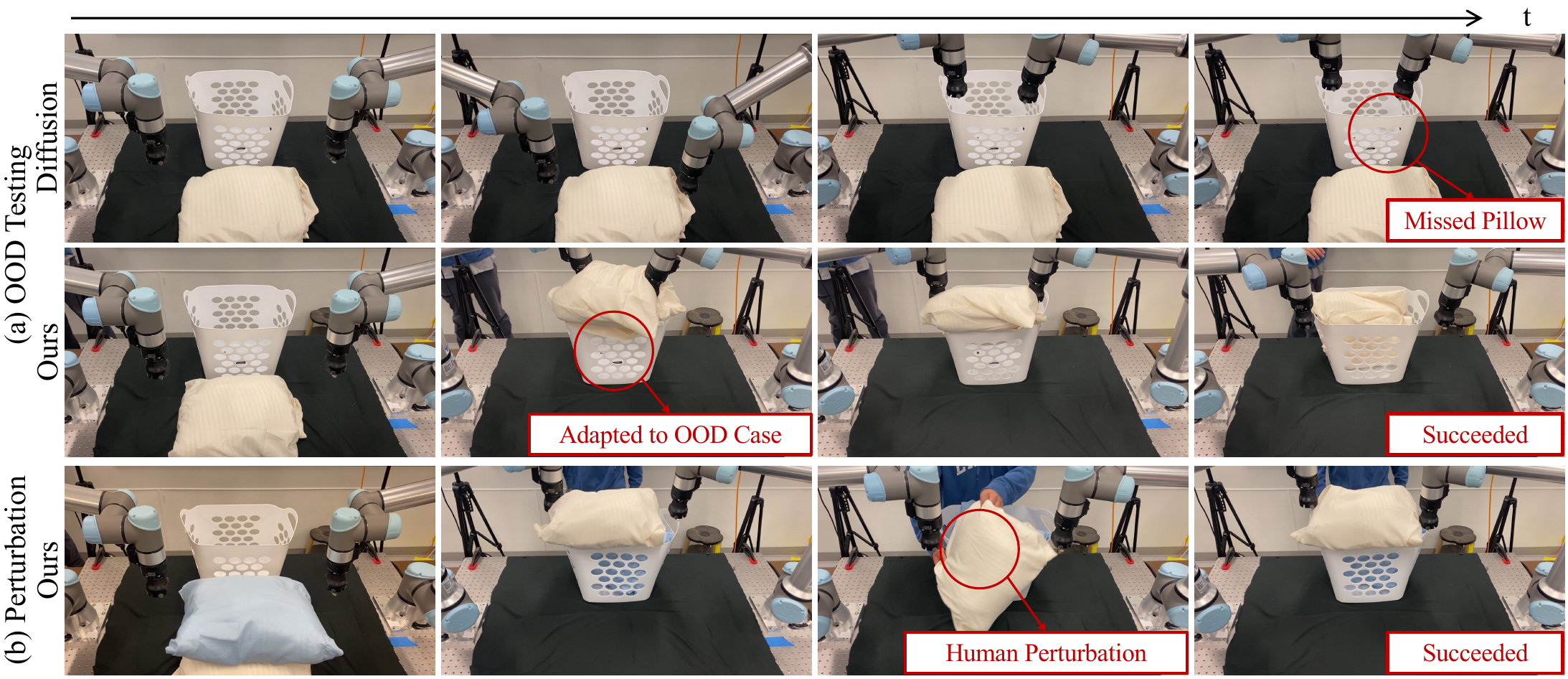}
  \caption{\textbf{Out-of-distribution (OOD), human perturbation testing, and emergent behavior of our policy in the laundry clean-up task.} (a) Unlike the training data, where two pillows are initially on the table, this out-of-distribution test started with only one pillow and an empty basket. Our method adapted and completed the task, while the baseline failed. (b) In a human perturbation scenario, after the robot placed two pillows in the basket, a human returned one pillow to the table. The robot successfully resumed and completed the task, demonstrating our method's robustness. }
  \label{fig:quali_ood}
\end{figure*}

\subsubsection{Bimanual Real-World Experiments}

\begin{table*}[tb]
    \begin{center}{
        \caption{\textbf{Runtime Cost Results on an Nvidia A40 GPU.} The runtime cost for each task, measured in seconds, is presented for different models.}
        \label{tab:runtime_cost}
        \begin{tabular}{l | c c c}
            \toprule
            \textbf{Model} & \textbf{Block Push (s)} & \textbf{Push-L (s)} & \textbf{Franka Kitchen (s)} \\
            \midrule
            Diffusion Policy & 0.590 ± 0.004 & 0.596 ± 0.001 & 0.604 ± 0.003 \\
            Improved Diffusion & 0.607 ± 0.017 & 0.602 ± 0.002 & 0.634 ± 0.029 \\
            {Ours} & \textbf{0.578 ± 0.007} & \textbf{0.588 ± 0.001} & \textbf{0.596 ± 0.007} \\
            \bottomrule
        \end{tabular}}
    \end{center}
\end{table*}

\begin{table*}[tb]
    \begin{center}{
        \caption{\textbf{Training Time Results for Different Models.} The training time for each task, measured in minutes, is presented for different models.}
        \label{tab:training_time}
        \begin{tabular}{l | c c c}
            \toprule
            \textbf{Model} & \textbf{Block Push (min)} & \textbf{Push-L (min)} & \textbf{Franka Kitchen (min)} \\
            \midrule
            Diffusion Policy & \textbf{188} & \textbf{56} & \textbf{246} \\
            Improved Diffusion & 255 & 86 & 307 \\
            {Ours} & 306 & 125 & 706 \\
            \bottomrule
        \end{tabular}}
    \end{center}
\end{table*}

\begin{itemize}
    \item \noindent\textbf{Qualitative Results:}
    Our real-world experiments showcased the qualitative performance of our method across multiple scenarios, demonstrating its robustness and adaptability. Figure \ref{fig:quali_real} depicts examples of the robot handling various manipulation tasks effectively. In the laundry cleanup task, our approach, which anticipates state changes, enabled the robot to manage the second pillow successfully. In contrast, the action-only model, which imitates the robot's movements, failed to account for the second pillow. Similarly, in the fruit holding task, our method managed to transport all four fruits, while the baseline approach resulted in one fruit being dropped. In the cluttered shelf picking task, our method succeeded in retrieving the pink book from a densely packed shelf without disturbing the objects above it, whereas the baseline method missed the book entirely.
    \item \noindent\textbf{Generalization and Robustness Testing:}
    To evaluate the generalization ability of our approach, we conducted out-of-distribution tests, including scenarios with human interventions and emergent behavior. Figure \ref{fig:quali_ood} illustrates the robot's performance in three challenging conditions. In the first condition, the initial setup differed from the training data, starting with only one pillow on the table instead of the usual two. Our method successfully adapted to this new condition and completed the task, unlike the baseline approach which failed. In the second condition, a human introduced a perturbation by placing a pillow back on the table after the robot had already placed it in the basket. Remarkably, the robot resumed and completed the task despite this interruption.

\end{itemize}


\subsection{Runtime and Training Time Analysis}

\subsubsection{Runtime Cost}
Table~\ref{tab:runtime_cost} presents the runtime cost results (in seconds) on an Nvidia A40 GPU. The table compares the runtime performance of three different models: Diffusion Policy, Improved Diffusion, and Ours. For each model, the runtime is measured across three tasks: Block Push, Push-L, and Franka Kitchen. The results show that our model achieves comparable inference times to the baselines while achieving superior task performance.

\subsubsection{Training Time}
Table~\ref{tab:training_time} summarizes the training time results (in minutes) for different models. Similar to the runtime cost analysis, the training time is evaluated for the Diffusion Policy, Improved Diffusion, and Our model across the three tasks: Block Push, Push-L, and Franka Kitchen. The results indicate that while our model requires longer training times, it achieves significantly better task performance.

\end{document}